%% file: bare_conf.tex
\documentclass[conference]{IEEEtran}
\ifCLASSINFOpdf
\else
\fi
\usepackage{graphicx}
\usepackage{multirow}
\usepackage{hhline}
\usepackage{color}
\newcommand\tab[1][.2cm]{\hspace*{#1}}

\begin{document}
%
\title{A Non-Parametric Learning Approach to Identify Online Human Trafficking}

\author{\IEEEauthorblockN{Hamidreza Alvari\\ and Paulo Shakarian}
\IEEEauthorblockA{Arizona State University\\
Tempe, Arizona\\
Email: \{halvari,shak\}@asu.edu}
\and
\IEEEauthorblockN{J.E. Kelly Snyder}
\IEEEauthorblockA{Find Me Group\\
Tempe, Arizona\\
Email: kelly@findmegroup.org}}


%


\maketitle

\begin{abstract}
Human trafficking is among the most challenging law enforcement problems which demands persistent fight against from all over the globe. In this study, we leverage readily available data from the website ``Backpage"-- used for classified advertisement-- to discern potential patterns of human trafficking activities which manifest online and identify most likely trafficking related advertisements. Due to the lack of ground truth, we rely on two human analysts --one human trafficking victim survivor and one from law enforcement, for hand-labeling the small portion of the crawled data. We then present a semi-supervised learning approach that is trained on the available labeled and unlabeled data and evaluated on unseen data with further verification of experts. 
\end{abstract}


%
\IEEEpeerreviewmaketitle

\section{Introduction}
Human trafficking has received increased national and societal concern over the past decade~\cite{HT_report2015}. According to the United Nation~\cite{unodc}, human trafficking is defined as the modern slavery or the trade of humans mostly for the purpose of sexual exploiting and forced labor, via different improper ways including force, fraud and deception. Human trafficking is among the challenging problems facing the law enforcement--it is difficult to identify victims and counter traffickers.  

Before the advent of the Internet, pimps were under the risks of being arrested by law enforcement, while advertising their victims on the streets~\cite{desplaces92}. However, the move to the Internet, has made it easier and less dangerous for both sex buyers and sellers, especially for the pimps~\cite{nicholas2012} as they no longer needed to advertise on the streets. There are now plethora of websites that host and provide sexual services, under categories of escort, adult entertainment, massage services, etc., which help pimps, traffickers and sex buyers (a.k.a. ``johns"), maintain their anonymity. Though some services such as Craiglist's adult section and myredbook.com were shut down recently, still there are many websites such as Backpage.com that provide such services and many new are frequently created. Traffickers even use dating and social networking websites, including Twitter, Facebook, Instagram and Tinder to reach out to the johns and their other followers. Although Internet has presented new trafficking related challenges for law enforcement, it has also provided readily and publicly available rich source of information which could be gleaned from online sex advertisements for fighting this crime~\cite{kennedy2012predictive}. 

Although, the Internet is being used for many other activities including attracting the victims, communicating with costumers and rating the escort services, here we only focus on the online advertisements. In this study, we use data crawled from the adult entertainment section of the website Backpage.com and propose a non-parametric learning approach to identify the most likely human trafficking related online advertisements out of the escort advertisements. To the best of our knowledge, this is the first study that employs both data mining and semi-supervised machine learning techniques to identify the potential human trafficking related advertisements given only a small portion of labeled data. We thus make the following contributions.

\begin{enumerate}
	\item We collected real posts from the U.S. cities represented on Backpage.com. The data was then preprocessed and cleaned.   
	\item Based on the literature, we created different groups of features that capture the characteristics of potential human trafficking activities. The less likely human trafficking related posts were then filtered out using these features.
	\item Due to the lack of ground truth, we relied on human analysts for hand-labeling small portion of the filtered data.
	\item We trained a semi-supervised learner on labeled and unlabeled data and sent back the identified highly human trafficking related advertisements to the experts for further verification. We then validated our approach on unseen data with further verification of experts.
\end{enumerate}

The rest of the paper is organized as follows. In Section II, we briefly provide the background of the problem of human trafficking. Next, we review the prior studies on human trafficking in Section III. Then in Section IV, we explain our data preparation and feature extraction scheme. Our unsupervised filtering and expert assisted labeling are explained in Sections V and VI, respectively. We detail our non-parametric learning approach in Sections VII. We conclude the paper by providing future research directions in Section VIII.   

\section{Background} 
The United States' Trafficking Victim Protection Act of 2000 (TVPA 2000)~\cite{tvpa2000}, was the first U.S. legislation passed against human trafficking. According to TVPA 2000, sex trafficking is a severe form of trafficking, where force, fraud or corecion are primary ways of inducing commercial sex act. Human Trafficking is a crime against humanity and is one of the most atrocious crimes of global magnitude. It is a \$150 billion industry of exploitation of children and young adults, utilizing humans for forced labor and sex trafficking worldwide. No country is immune and the problem is rapidly growing with little to no law enforcement addressing the issue and approximately 161 countries affected. Human trafficking is considered to be a form of modern day slavery. Humans are controlled, exploited, abused,  forced into prostitution and labor of servitude in some form and all under the threat of punishment if they do not perform their required duties.

The Find Me Group (FMG) was founded by retired DEA Special Agent Jerry ``Kelly" Snyder in 2002 primarily to locate missing persons. The natural evolution of the group in locating missing persons was to allocate resources for locating victims in human trafficking, as well as identifying the persons responsible and reporting these organizations to law enforcement. The FMG consists of current and retired law enforcement agents and officers with a wide-range of investigative expertise, including but not limited to linguistics, handwriting analysis, body language, missing persons and homicide. The search and rescue component of the FMG is also comprised of current and retired law enforcement officers and agents with 28 years of field management skills in locating missing persons. The FMG has an additional advantage by using trained experts/sources that provide detailed location information of human trafficking victims.  

The ultimate goal of the current project is to identify missing persons which are connected to human trafficking organizations. This can be done by identifying their locations, utilizing logistical methodology with an additional focus on their financial status and reporting assets to worldwide law enforcement.

\section{Related Work}
Recently, several studies have examined the role of the Internet and related technology in facilitating human trafficking\cite{hughes2005demand,hughes2002use,latonero2011human}. For example, the work of~\cite{hughes2005demand} studied how closely sex trafficking is intertwined with new technologies. According to~\cite{hughes2002use}, ``The sexual exploitation of women and children is a global human rights crisis that is being escalated by the use of new technologies". Researchers have studied the relationship between new technologies and human trafficking and advantages of the Internet for sex traffickers. For instance, according to~\cite{latonero2011human}, findings from a group of experts from the Council of Europe demonstrate that the Internet and sex industry are closely interlinked and the volume and content of the material on the Internet promoting human trafficking are unprecedented.

One of the earliest works which leveraged data mining techniques for online human trafficking was~\cite{latonero2011human}, where the authors conducted an analysis of data on the adult section of the website Backpage.com. Their findings confirmed that the female escort post frequency would increase in Dallas, Texas, leading up to Super Bowl 2011 event. In a similar attempt, other studies~\cite{dominique2015,2016arXiv160205048M} have investigated the impact of large public events such as Super Bowl on sex trafficking by exploring advertisement volume, trends and movement of advertisements along with the scope and volume of demand associated with such events. The work of~\cite{dominique2015}, for instance, concludes that in large events like Super Bowl which attract significant amount of concentration of people in a relatively short period of time and in a confined urban area, could be a desirable location for sex traffickers to bring their victims for commercial sexual exploitation. Similarly, the data-driven approach of~\cite{2016arXiv160205048M} shows that in some but not all events, one can see a correlation between the occurrence of the event and statistically significant evidence of an influx of sex trafficking activity. Also, certain studies~\cite{conf/semweb/SzekelyKSPSYKNM15} have tried to build large distributed systems to store and process the available online human trafficking data in order to perform entity resolution and create ontological relations between the entities. 

Beyond these works, the work of~\cite{2015arXiv150906659N}, studied the problem of isolating sources of human trafficking from online advertisements with a pairwise entity resolution approach. Specifically, they trained a classifier to predict if two ads are from the same source, using phone numbers as a strong feature. Then, this classifier was used to perform entity resolution using a heuristically learned value for the score of classifier. Another work of~\cite{kennedy2012predictive} used Backpage.com data and extracted most likely human trafficking spatio-temporal patterns with the help of law enforcement. Note that unlike our method, this work did not employ any machine learning methodologies for automatically identifying the human trafficking related advertisements. 
The work of~\cite{doi:10.1080/23322705.2015.1015342} also used machine learning techniques by training a supervised learning classifier on labeled data (based on the phone numbers of known traffickers) provided by a victim advocacy group, for the ad-classification problem. We note that while phone numbers can provide very precise set of positive labeled data, there  are clearly many posts with previously unseen phone numbers. In contrast, we do not solely rely on the phone numbers for labeling the data. Instead, our experts analyze the whole post's content to identify whether it is human trafficking related or not. Indeed, we first filter out the most likely advertisements using several feature groups and pass a small sample to the experts for hand-labeling. Then, we train semi-supervised learner on both labeled and unlabeled data which in turn let us evaluate our approach on the new coming (unseen) data as well. We note that our semi-supervised approach can also be used as a complementary method to procedures such as those described in~\cite{doi:10.1080/23322705.2015.1015342} as we can significantly expand the training set for use with supervised learning.

\section{Data Collection Effort}
\begin{figure}[!ht]
	\caption{A real post from Backpage.}
	\centering
	\includegraphics[width=0.45\textwidth]{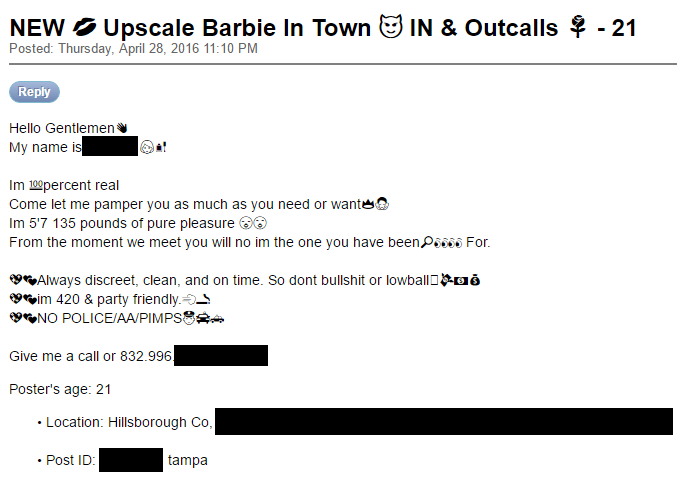}
	\label{fig:post1}
\end{figure}

\begin{figure}[!ht]
	\caption{Phone distribution by different states.}
	\centering
	\includegraphics[width=0.5\textwidth]{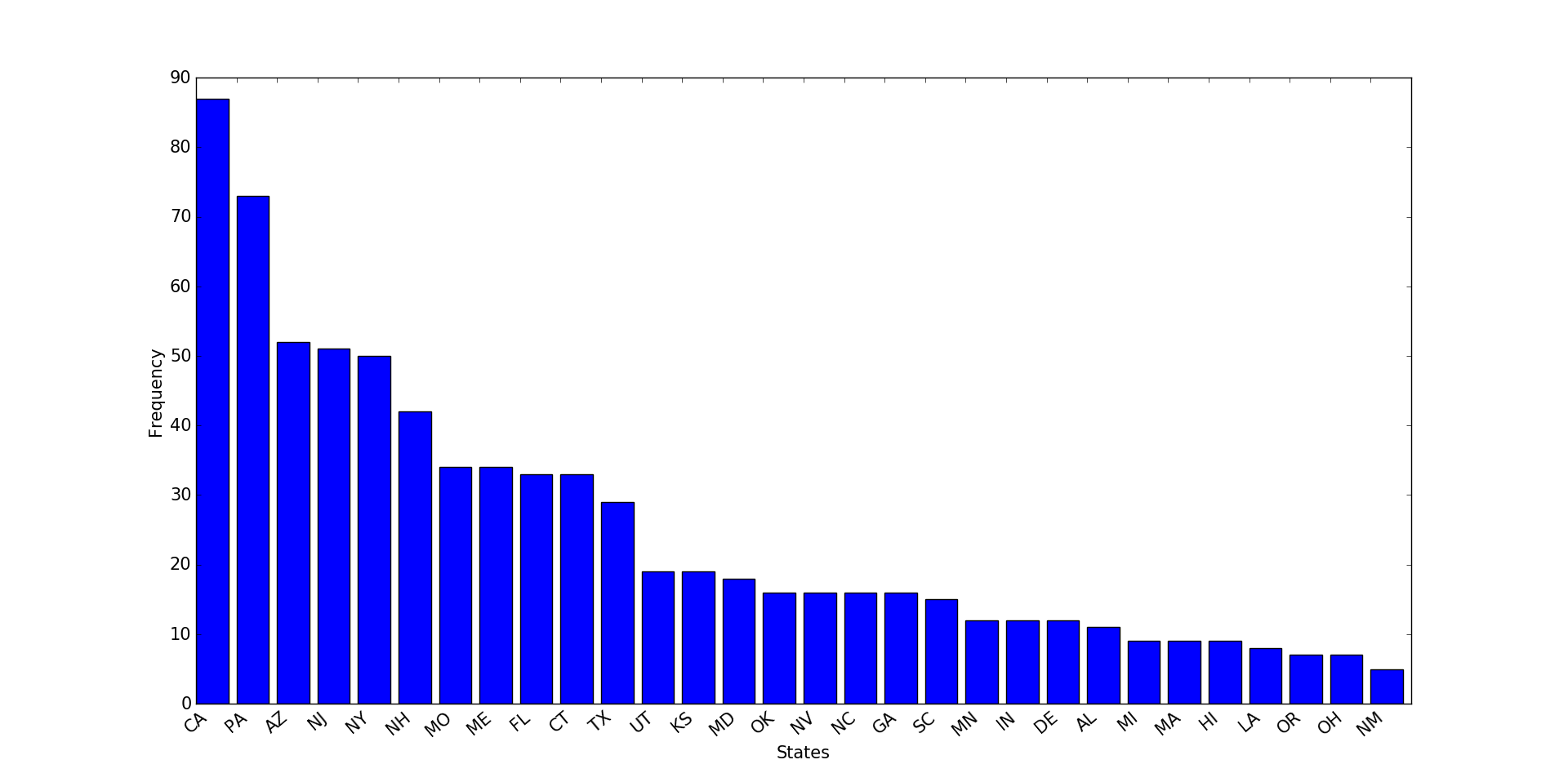}
	\label{fig:dist}
\end{figure}

We collected about 20K publicly available listings from the U.S. posted on Backpage.com in March, 2016. Each post includes a title, description, time stamp, the poster's age, poster's ID, location, image, video and sometimes audio. The description usually lists the attributes of the individual(s) and contact phone numbers. In this work we only focus on the textual component of the data. This free-text data required significant cleaning due to a variety of issues common to textual analytics (i.e. misspellings, format of phone numbers, etc.). We also acknowledge that the information in data could be intentionally inaccurate, such as the poster's name, age and even physical appearance (i.e. bra cup size, weight). Figure~\ref{fig:post1} shows an actual post from Backpage.com. To illustrate the geographic diversity of the listings, we also plot the phone distribution with respect to the different states in Figure~\ref{fig:dist}. Note that for brevity, we only show those with a frequency greater than 5. 

Next, we explain the most important characteristics of potential human trafficking advertisements captured by our feature groups. 

\subsection{Feature Engineering}
\begin{figure}[!ht]
	\caption{An evidence of human trafficking. The boxes and numbers in red, indicate the features and their corresponding group numbers (see also Table~\ref{tb:feat}).}
	\centering
	\includegraphics[width=0.45\textwidth]{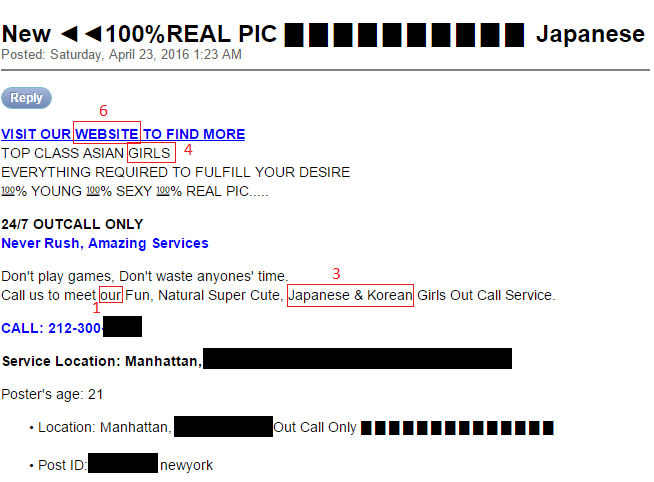}
	\label{fig:post2}
\end{figure}

Though many advertisements on Backpage.com are posted by posters selling their own services without coercion and intervention of traffickers, some do exhibit many common trafficking triggers. For example, in contrast to the previous advertisements, Figure~\ref{fig:post2} shows an advertisement that could be an evidence of human trafficking. This advertisement has several potential properties of human trafficking including advertising for multiple escorts with the first individual coming from Asia and very young. In the followings, we discuss such common properties of human trafficking related advertisements, in more details.

Inspired from literature, we define and extract 6 groups of features from advertisements, shown in Table~\ref{tb:feat}, which could be amongst the strong indicators of the human trafficking. In what follows, we briefly describe each group of features used in our work. Each feature listed is treated as a binary variable.

\begin{table}[t]
	\centering
	\caption{Different groups of features used in our work.}
	\begin{tabular}{|c|l|c|} 
		\hline
		\textbf{No.} & \textbf{Feature Group} & \textbf{Ref.}\\ [0.5ex] 
		\hline\hline
		1 & Advertisement Language Pattern & \cite{kennedy2012predictive,Li:2008:IKC:1478784,conf/setn/KanarisKS06}\\  
		2 & Words and Phrases of Interest & \cite{hetter2012,Lloyd2012,Goodman2011} \\
		3 & Countries of Interest& \cite{HT_report2015}\\
		4 & Multiple Victims Advertised & \cite{kennedy2012predictive}\\
		5 & Victim Weight & \cite{tvpa2000,weightchart}\\
		6 & Reference to Website or Spa Massage Therapy & \cite{kennedy2012predictive}\\
		\hline
	\end{tabular}
	
	\label{tb:feat}
\end{table}

\subsubsection{\textbf{Advertisement Language Pattern}} The first group consists of different language related features. For the first and second features, we identify posts which has third person language (more likely to be written by someone other than the escort) and posts which contain first person plural pronouns such as `we' and `our' (more likely to be an organization)~\cite{kennedy2012predictive}.

To ensure their anonymity, traffickers would deploy techniques to generate diverse information and hence make their posts look more complicated. They usually do this to avoid being identified by either human analyst or automated programs. Thus, to obtain the third feature, we take an approach from complexity theory, namely \textit{Kolmogorov complexity} which is defined as the length of the shortest program to reproduce the advertisement content on a universal machine such as Turing Machine~\cite{Li:2008:IKC:1478784}. We approximate the Kolmogorov complexity of an advertisement's content, by simply computing the Entropy of the content~\cite{Li:2008:IKC:1478784} as follows. Let $X$ denote the content and $x_i$ be a given word in the content. We use the following equation~\cite{shannon2001mathematical} to calculate the Entropy of the content.
\begin{equation}
H(X) = -\sum_{i=1}^n{P(x_i)\log_2 P(x_i)}
\end{equation}

We expect higher values of the Entropy correspond to human trafficking. Finally, we discretize the result by using the threshold of 4 which was found empirically in our experiments.

Next, we use word-level \textit{n}-grams to find the common language patterns of the advertisements, as the character-level \textit{n}-grams have already shown to be useful in detecting unwanted content for Spam detection~\cite{conf/setn/KanarisKS06}. We set $n=4$ and compute the normalized \textit{n}-grams (using TF-IDF) of the advertisement's content and use threshold of 0.5 to binarize their values. This gives us 6 more features to include into our feature set. Overall, we have 9 features related to the language of the advertisement.\\ 

\subsubsection{\textbf{Words and Phrases of Interest}} Despite the fact that advertisements on Backpage.com do not directly mention sex with children, costumers who prefer children, know to look for words and phrases such as ``\textit{sweet}, \textit{candy}, \textit{fresh}, \textit{new in town}, \textit{new to the game}"~\cite{hetter2012,Lloyd2012,Goodman2011}. We thus investigate within the posts to see if they contain such words as they could be highly related with human trafficking in general.\\

\subsubsection{\textbf{Countries of Interest}} We identify if the individual being escorted is coming from other countries such as those in Southeast Asia (especially from China, Vietnam, Korea and Thailand, as we observed in our data)~\cite{HT_report2015}.\\

\subsubsection{\textbf{Multiple Victims Advertised}} Some advertisements advertise for multiple women at the same time. We consider the presence of more than one girl as a potential evidence of organized human trafficking~\cite{kennedy2012predictive}.\\

\subsubsection{\textbf{Victim Weight}} We take into account weight of the individual being escorted as a feature (if it is available). This information is particularly useful assuming that for the most part, lower body weights (under 110 lbs) correlate with smaller and underage girls~\cite{tvpa2000,weightchart} and thereby human trafficking.\\ 

\subsubsection{\textbf{Reference to Website or Spa Massage Therapy}} The presence of a link in the advertisement, either referencing to an outside website (especially infamous ones) or spa massage therapy, could be an indicator of more elaborate organization~\cite{kennedy2012predictive}. In case of spa therapy, we observed many advertisements interrelated with advertising for young Asian girls and their erotic massage abilities. Therefore, the last group has two binary features for the presence of both website and spa.\\

In order to extract these features, we first clean the original data and conduct preprocessing. Then we draw 999 instances out of our dataset for further analysis, as they might be evidence of human trafficking-- this is described in the next section. 

\section{Unsupervised Filtering}
Having detailed our feature set, we now construct feature vectors for each instance by creating a vector of 15 binary features that correspond to the important characteristics of human trafficking related posts.

\begin{figure}[!ht]
	\caption{Two clusters of a portion of the filtered data set combined with random samples from the remainder of the samples, in the trasformed feature space.}
	\centering
	\includegraphics[width=0.4\textwidth]{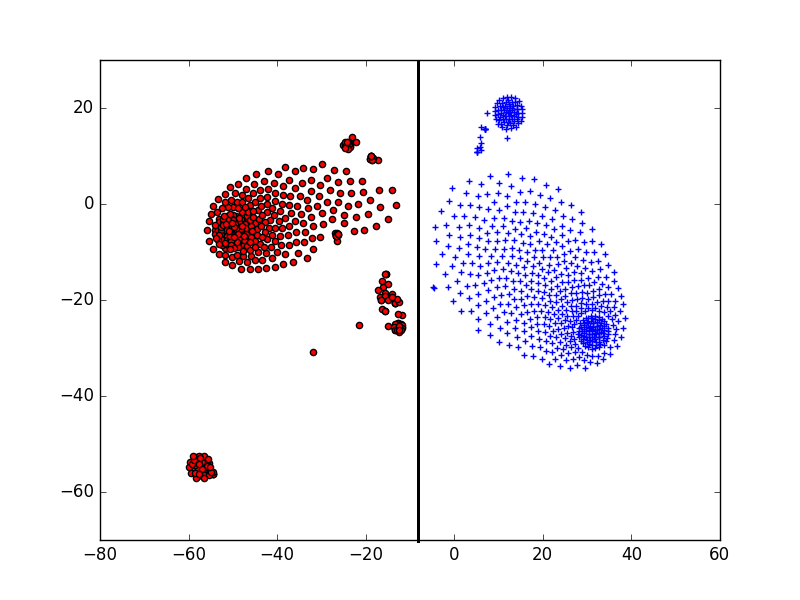}
	\label{fig:vis}
\end{figure}

We obtain 999 instances from our dataset by filtering out samples that do not posses any of the binary features. We will refer to this as our \textit{filtered} dataset. In Figure~\ref{fig:vis}, we visualize 500 of the 999 samples and an additional 500 samples outside of the filtered dataset (i.e., from the remainder of the samples) we studied in a 2-D projection (using the t-SNE transformation~\cite{maaten2008visualizing}).  We clustered the visualized samples into 2 clusters (using K-means) and found the clusters to be purely either inside or outside of the sampled data (100\% of samples in cluster 1 were from the identified listings and 100\% of samples in cluster 2 were from outside this group).

We then create a second feature space that is used through the remainder of the paper. Using Latent Drichlet Allocation (LDA)~\cite{blei2003latent} topic modeling from Python package \textit{gensim}~\cite{rehurek_lrec}, we identify 25 most representative topics out of the filtered dataset. This allows us to uncover the hidden thematic structure in the data. Further, we rely on the document-topic distribution given by the LDA (here each document is seen as a mixture of topics) to distinguish the normal advertisements (outliers) from highly human trafficking related ones. More specifically, we treat each listing in the filtered data as a vector of 25 probabilistic values provided by LDA's document-topic distribution-- this feature space is used in the next step. 
	

Moreover, since we lack ground truth for our data, we rely on human analysts (experts) for labeling the listings as either human trafficking or not. In the next section, we select a smaller yet finer grain subset of this data to be sent to the experts. This alleviates the burden of the tedious work of hand-labeling.
\section{Expert Assisted Labeling}




We first obtain a sample of 150 listings from the filtered dataset. This set of listings was labeled by two human experts: a previous human trafficking victim and a law enforcement officer who specialized in this type of crime. From this subset, a law enforcement professional and human trafficking victim identified 38 and 139 instances (respectively) to be human trafficking related instances. Among them, there were 31 records for which both experts agreed were highly related to human trafficking. Thus, we now have 31 confirmed positive samples, but still have large amounts of unlabeled examples (849 instances) in our dataset. We summarize the data statistics in Table~\ref{tb:st}. Any sample for which at least one expert labeled as negative, we treated as a negative sample.
\begin{table}[h!]
	\centering
	\caption{Description of the dataset.}
	\begin{tabular}{|l|c|c|c|c|}
		\cline{1-5}
		\textbf{Name}          & \multicolumn{4}{c|}{\textbf{Value}}\\
		\hhline{=====}
		Raw      & \multicolumn{4}{c|}{20,822}  \\ \cline{1-5}
		Filtered & \multicolumn{4}{c|}{999} \\ \cline{1-5}
		Unlabeled & \multicolumn{4}{c|}{849} \\ \cline{1-5}
		Labeled  & Expert 1 & Expert 2 & Intersection & Union \\ \cline{2-5}
		\tab{Positive} & 38  & 139 & \underline{31} & 146\\
		\tab{Negative} & 112   & 11 & 4 & \underline{119}\\ \cline{1-5}
	\end{tabular}
	\label{tb:st}
\end{table}

In the next section, we explain how we deploy a non-parametric learning approach to identify the labels of the rest of the data to be sent for further expert verification.   
\section{Non-Parametric Learning}
We use the Python package \textit{scikit-learn}~\cite{scikit-learn} for training semi-supervised learner on the filtered dataset. There are two label propagation semi-supervised (non-parametric) based models in this package, namely, LabelPropagation and LabelSpreading~\cite{chapelle2006}. These models rely on the geometry of the data induced by both labeled and unlabeled instances as opposed to the supervised models which only use the labeled data~\cite{chapelle2006}. This geometry is usually represented by a graph $G=(V,E)$, with the nodes $V$ represent the training data and edges $E$ represent the similarity between them~\cite{chapelle2006} in the form of weight matrix $\textbf{W}$. Given the graph $G$, a basic approach for semi-supervised learning is through propagating labels on the graph~\cite{chapelle2006}. Due to the higher performance achieved, we chose to use LabelSpreading model. We conducted experiment with the two built-in kernels radial basis function (RBF) and K-nearest neighbor (KNN) in label propagation models and report the results in Table~\ref{tb:pr}. Note that we only reported the precision when 119 negative samples (labeled by either of the experts) were used in the learning process. We did so because of the reasonable number of the positive labels assigned by either of the kernels in presence of these negative instances (our experts had limited time to validate the labels of the data).

\begin{table}[h!]
	\centering
	\caption{Validated results on unlabeled data for both kernels.}
	\begin{tabular}{|l|c|c|c|c|}
		\cline{1-5}
		\textbf{Name}          & \multicolumn{4}{c|}{\textbf{Value}}\\
		\hhline{=====}
		  & Positive & Negative & Positive & Precision \\ 
		Kernel & (Learner) & (Learner) & (Experts) & (Positive) \\
		\cline{2-5}
		\tab{RBF (Union)}  & \underline{145} & 704 & 134 & 92.41\% \\
		\tab{RBF (Intersection)}  & 848 & 1 & - & - \\ 
		\tab{KNN (Union)}  & \underline{188} & 661 & 170 & 90.42\% \\ 
		\tab{KNN (Intersection)}  & 849 & 0 & -  & - \\ \cline{1-5}
	\end{tabular}
	\label{tb:pr}
\end{table}

\begin{figure}[!ht]
	\caption{Word cloud for the positive instances assigned by RBF.}
	\centering
	\includegraphics[width=0.4\textwidth]{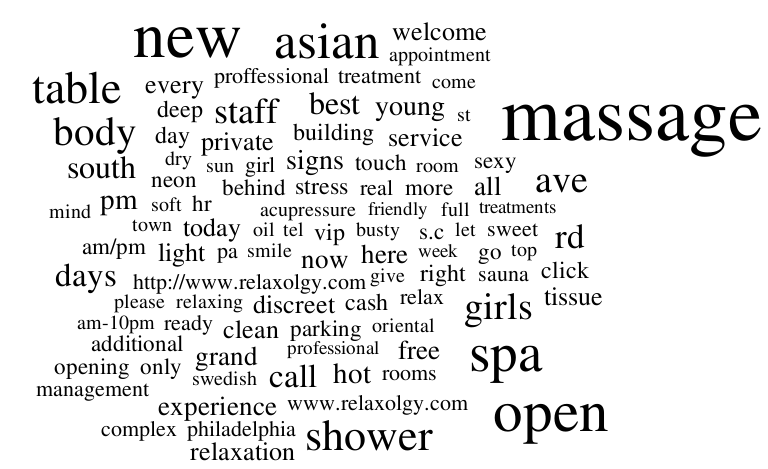}
	\label{fig:cloud1}
\end{figure}

\begin{figure}[!ht]
	\caption{Word cloud for the positive instances assigned by KNN.}
	\centering
	\includegraphics[width=0.4\textwidth]{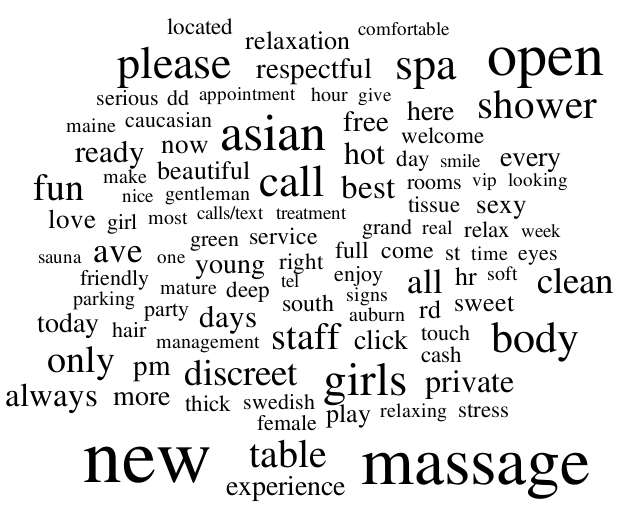}
	\label{fig:cloud2}
\end{figure}

As we see from this table, out of 849 unlabeled data, our learner with RBF and KNN kernels assigned positive labels to the 145 and 188 instances, respectively. Next, we pass the identified positive labels to the experts for further verification. Our approach with RBF and KNN correctly identified 134 and 170 labels out of 145 and 188 positive instances and achieved precision of 92.41\% and 90.42\%, respectively. We further demonstrate the word clouds for the positive instances assigned by RBF and KNN, in Figure~\ref{fig:cloud1} and Figure~\ref{fig:cloud2}, respectively.

\section{Conclusion}
Readily available online data from escort advertisements could be leveraged in favor of fighting against the human trafficking. In this study, having focused on textual information from available data crawled from Backpage.com, we identified if an escort advertisement can be reflective of human trafficking activities. More specifically, we first propose an unsupervised filtering approach to filter out the data which are more likely involved in trafficking. We then trained a semi-supervised learner on small portion of such data, hand-labeled by human trafficking experts, to identify the labels for unseen data. The results suggest our non-parametric approach is successful at identifying the potential human trafficking related advertisements.

In future work we seek to extract the underlying network of the data to find interesting patterns such as the most influential nodes as they might indicate the known pimps and traffickers. We would also like to replicate the study by integrating more features, especially those supported by the criminology literature.


\section*{Acknowledgment}
This work was funded by the Find Me Group, a 501(c)3 dedicated to bring resolution and closure to families of missing persons.  https://www.findmegroup.org/



%



\input{ref.bbl}

\end{document}

%% file: ref.bbl

%% file: bare_conf.bbl
\begin{thebibliography}{10}
\providecommand{\url}[1]{#1}
\csname url@samestyle\endcsname
\providecommand{\newblock}{\relax}
\providecommand{\bibinfo}[2]{#2}
\providecommand{\BIBentrySTDinterwordspacing}{\spaceskip=0pt\relax}
\providecommand{\BIBentryALTinterwordstretchfactor}{4}
\providecommand{\BIBentryALTinterwordspacing}{\spaceskip=\fontdimen2\font plus
\BIBentryALTinterwordstretchfactor\fontdimen3\font minus
  \fontdimen4\font\relax}
\providecommand{\BIBforeignlanguage}[2]{{%
\expandafter\ifx\csname l@#1\endcsname\relax
\typeout{** WARNING: IEEEtran.bst: No hyphenation pattern has been}%
\typeout{** loaded for the language `#1'. Using the pattern for}%
\typeout{** the default language instead.}%
\else
\language=\csname l@#1\endcsname
\fi
#2}}
\providecommand{\BIBdecl}{\relax}
\BIBdecl

\bibitem{HT_report2015}
``{Trafficking in Persons Report},'' July 2015.

\bibitem{unodc}
``{UNODC on human trafficking and migrant smuggling},'' 2011.

\bibitem{desplaces92}
C.~Desplaces, ``{Police Run `Prostitution' Sting; 19 Men Arrested, Charged in
  Fourth East Dallas Operation.}'' Nov 1992.

\bibitem{nicholas2012}
K.~Nicholas~D., ``{How Pimps Use the Web to Sell Girls.}'' Jan 2012.

\bibitem{kennedy2012predictive}
E.~Kennedy, ``{Predictive patterns of sex trafficking online},'' 2012.

\bibitem{tvpa2000}
``{Trafficking Victims Protection Act of 2000},'' 2000.

\bibitem{hughes2005demand}
D.~M. Hughes \emph{et~al.}, ``{The demand for victims of sex trafficking},''
  \emph{Women’s Studies Program, University of Rhode Island}, 2005.

\bibitem{hughes2002use}
D.~M. Hughes, ``{The Use of New Communications and Information Technologies for
  Sexual Exploitation of Women and Children},'' \emph{Hastings Women's Law
  Journal}, vol.~13, no.~1, pp. 129--148, 2002.

\bibitem{latonero2011human}
M.~Latonero, ``{Human trafficking online: The role of social networking sites
  and online classifieds},'' \emph{Available at SSRN 2045851}, 2011.

\bibitem{dominique2015}
D.~{Roe-­‐Sepowitz}, J.~{Gallagher}, K.~{Bracy}, L.~{Cantelme},
  A.~{Bayless}, J.~{Larkin}, A.~{Reese}, and L.~{Allbee}, ``{Exploring the
  Impact of the Super Bowl on Sex Trafficking},'' Feb. 2015.

\bibitem{2016arXiv160205048M}
K.~{Miller}, E.~{Kennedy}, and A.~{Dubrawski}, ``{Do Public Events Affect Sex
  Trafficking Activity?}'' \emph{ArXiv e-prints}, Feb. 2016.

\bibitem{conf/semweb/SzekelyKSPSYKNM15}
P.~A. Szekely, C.~A. Knoblock, J.~Slepicka, A.~Philpot, A.~Singh, C.~Yin,
  D.~Kapoor, P.~Natarajan, D.~Marcu, K.~Knight, D.~Stallard, S.~S.
  Karunamoorthy, R.~Bojanapalli, S.~Minton, B.~Amanatullah, T.~Hughes,
  M.~Tamayo, D.~Flynt, R.~Artiss, S.-F. Chang, T.~Chen, G.~Hiebel, and
  L.~Ferreira, ``{Building and Using a Knowledge Graph to Combat Human
  Trafficking.}'' in \emph{International Semantic Web Conference (2)}, ser.
  Lecture Notes in Computer Science, vol. 9367.\hskip 1em plus 0.5em minus
  0.4em\relax Springer, 2015, pp. 205--221.

\bibitem{2015arXiv150906659N}
C.~{Nagpal}, K.~{Miller}, B.~{Boecking}, and A.~{Dubrawski}, ``{An Entity
  Resolution approach to isolate instances of Human Trafficking online},''
  \emph{ArXiv e-prints}, Sep. 2015.

\bibitem{doi:10.1080/23322705.2015.1015342}
A.~Dubrawski, K.~Miller, M.~Barnes, B.~Boecking, and E.~Kennedy, ``Leveraging
  publicly available data to discern patterns of human-trafficking activity,''
  \emph{Journal of Human Trafficking}, vol.~1, no.~1, pp. 65--85, 2015.

\bibitem{Li:2008:IKC:1478784}
M.~Li and P.~M. Vitnyi, \emph{{An Introduction to Kolmogorov Complexity and Its
  Applications}}, 3rd~ed.\hskip 1em plus 0.5em minus 0.4em\relax Springer
  Publishing Company, Incorporated, 2008.

\bibitem{conf/setn/KanarisKS06}
I.~Kanaris, K.~Kanaris, and E.~Stamatatos, ``{Spam Detection Using Character
  N-Grams.}'' in \emph{SETN}, ser. Lecture Notes in Computer Science, vol.
  3955.\hskip 1em plus 0.5em minus 0.4em\relax Springer, 2006, pp. 95--104.

\bibitem{hetter2012}
K.~Hetter, ``{Fighting sex trafficking in hotels, one room at a time},'' March
  2012.

\bibitem{Lloyd2012}
R.~Lloyd, ``{An Open Letter to Jim Buckmaster},'' April 2012.

\bibitem{Goodman2011}
J.~{Dickinson Goodman} and M.~{Holmes}, ``{Can We Use RSS to Catch Rapists},''
  2011.

\bibitem{weightchart}
\BIBentryALTinterwordspacing
``{Average Height to Weight Chart - Babies to Teenagers}.'' [Online].
  Available:
  \url{http://www.disabled-world.com/artman/publish/height-weight-teens.shtml}
\BIBentrySTDinterwordspacing

\bibitem{shannon2001mathematical}
C.~E. Shannon, ``A mathematical theory of communication,'' \emph{ACM SIGMOBILE
  Mobile Computing and Communications Review}, vol.~5, no.~1, pp. 3--55, 2001.

\bibitem{maaten2008visualizing}
L.~van~der Maaten and G.~Hinton, ``{Visualizing High-Dimensional Data Using
  t-SNE},'' 2008.

\bibitem{blei2003latent}
D.~M. Blei, A.~Y. Ng, and M.~I. Jordan, ``Latent dirichlet allocation,''
  \emph{the Journal of machine Learning research}, vol.~3, pp. 993--1022, 2003.

\bibitem{rehurek_lrec}
R.~{\v R}eh{\r u}{\v r}ek and P.~Sojka,
  ``\BIBforeignlanguage{English}{{Software Framework for Topic Modelling with
  Large Corpora}},'' in \emph{\BIBforeignlanguage{English}{{Proceedings of the
  LREC 2010 Workshop on New Challenges for NLP Frameworks}}}.\hskip 1em plus
  0.5em minus 0.4em\relax Valletta, Malta: ELRA, May 2010, pp. 45--50,
  \url{http://is.muni.cz/publication/884893/en}.

\bibitem{scikit-learn}
F.~Pedregosa, G.~Varoquaux, A.~Gramfort, V.~Michel, B.~Thirion, O.~Grisel,
  M.~Blondel, P.~Prettenhofer, R.~Weiss, V.~Dubourg, J.~Vanderplas, A.~Passos,
  D.~Cournapeau, M.~Brucher, M.~Perrot, and E.~Duchesnay, ``Scikit-learn:
  Machine learning in {P}ython,'' \emph{Journal of Machine Learning Research},
  vol.~12, pp. 2825--2830, 2011.

\bibitem{chapelle2006}
Y.~{Bengio}, O.~{Delalleau}, and N.~{Le Roux}, \emph{{In Semi-Supervised
  Learning}}.\hskip 1em plus 0.5em minus 0.4em\relax MIT Press, 2006.

\end{thebibliography}
